\documentclass{article}

\usepackage[T1]{fontenc}
\PassOptionsToPackage{sort&compress}{natbib}  
\usepackage[final]{corl_2026} 
\makeatletter
\renewcommand{\@noticestring}{CoRL 2026 Workshop on Learning from Corrections and Interventions, Austin, Texas, USA.}
\makeatother
\hypersetup{pdftitle={What Can One Sentence Fix? A Failure Taxonomy and a Minimal Language Correction for Frozen VLA Policies}, pdfauthor={Owen Kwon, Pablo Ortega-Kral, Arthur Bucker, Jean Oh}, pdfsubject={CoRL 2026 Workshop on Learning from Corrections and Interventions}}

\usepackage{amsmath,amssymb}
\usepackage{booktabs}
\usepackage{graphicx}
\usepackage{multirow}
\usepackage{tabularx}
\usepackage{xcolor}
\usepackage{enumitem}
\usepackage{needspace}
\newcolumntype{L}{>{\raggedright\arraybackslash}X}
\newcolumntype{C}{>{\centering\arraybackslash}X}

\newcommand{\corr}{\textsc{CorrectVLA}}

\title{What Can One Sentence Fix?\\A Failure Taxonomy and a Minimal Language\\Correction for Frozen VLA Policies}

\author{Owen Kwon$^{1}$, Pablo Ortega-Kral$^{1}$, Arthur Bucker$^{1}$, and Jean Oh$^{1}$\\
$^{1}$Carnegie Mellon University}

\begin{document}
\maketitle

\begin{abstract}
When a deployed vision-language-action (VLA) policy fails, fine-tuning it again for every new setting is costly, yet the person watching it can usually say what went wrong in one sentence. We ask what such a sentence can and cannot fix in a frozen policy. Labelling every failing LIBERO-90 task (out of distribution) for OpenVLA-OFT and $\pi_{0.5}$, we find that the failures sort into four modes and that only one, \emph{execution misalignment}, where the policy reaches the right target but miscalibrates how far or how hard to move, is within reach of a sentence; the other three lack a target or a plan, not a magnitude. For this mode we define the narrowest human correction: one retrospective sentence per task, grounded by an LLM into an action dimension, direction, magnitude term, and time window, becomes a short bias on the frozen policy's actions, after which the policy completes the rest on its own, episode after episode. Re-executed on its task's failed rollouts, it recovers 65 of all 139 in-distribution and 33 of all 328 out-of-distribution failures in simulation; on a real xArm7 after the robot base is moved, it raises pick-and-place from 10/30 to 30/30 in each of two same-object conditions and from 2/30 to 24/30 with a new object, while two long-horizon tasks stay at 0/30 with and without correction, as the taxonomy predicts. A naively prompted VLM that watches the same failures and writes the same kind of correction recovers none in up to 10 attempts, suggesting that the value lies in the person's calibrated sense of magnitude and timing rather than in language itself. Videos: \url{https://correctvla.github.io/}
\end{abstract}

\keywords{Learning from corrections, VLA, Failure taxonomy}

\section{Introduction}
\label{sec:intro}

Vision-language-action (VLA) policies~\citep{openvla,pi0,pi05,octo,rt2} execute manipulation tasks from a sentence of instruction and carry that competence into new scenes. Yet under a new camera pose or table height a VLA often keeps the task structure (approach, grasp, transport) intact and fails on a magnitude: the grasp is a few centimeters shallow, the push a little weak. The durable recourse is still to collect demonstrations and fine-tune~\citep{pi0,oft} in each new setting; test-time steering keeps the weights frozen but biases or selects the policy's own samples~\citep{itps,forewarn}, so it cannot realize an intent the policy could not already generate. Language can also reshape a given trajectory~\citep{latte,extract}, but a VLA replans in closed loop. Which failures a person could fix, and how to directly steer a pretrained policy out of them, remain unclear.

\begin{figure}[t]
  \centering
  \includegraphics[width=\linewidth]{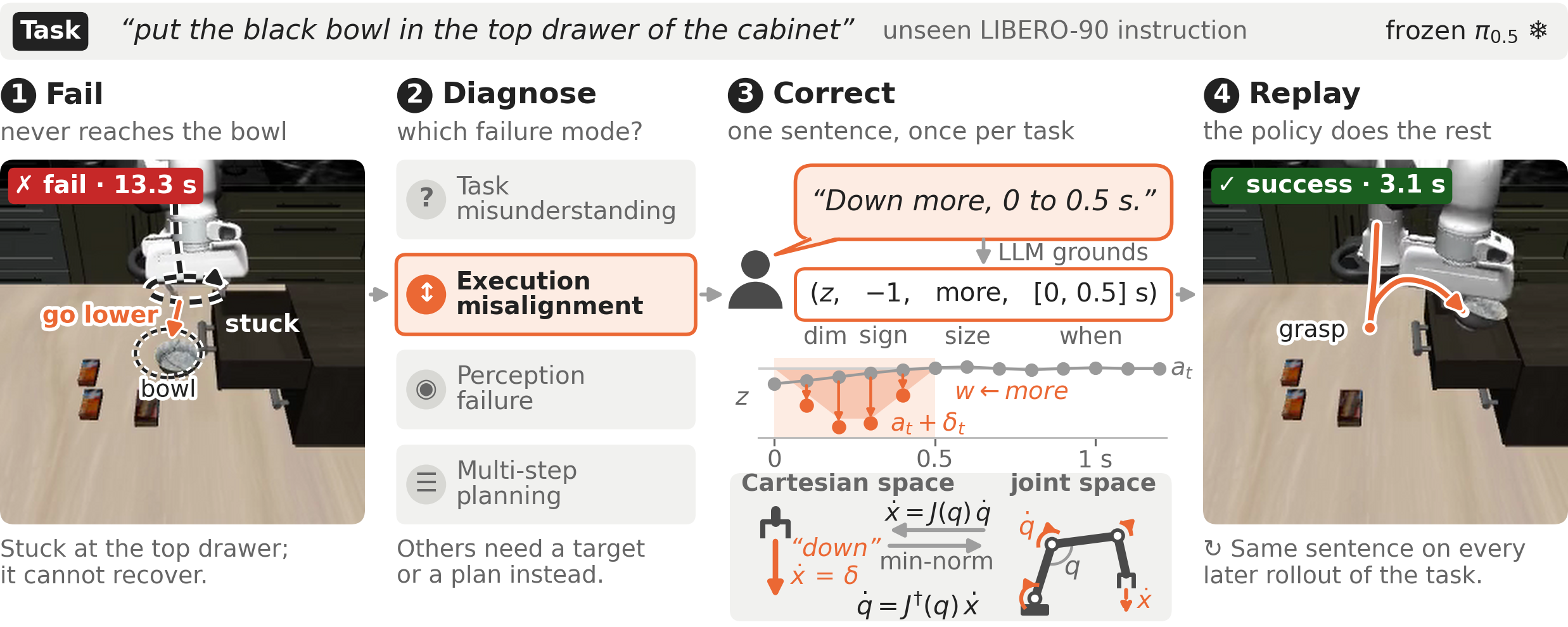}
  \caption{\textbf{Overview.} The frozen $\pi_{0.5}$ gets stuck at the top drawer and cannot recover (1), an execution misalignment (2). One sentence becomes a short bias $\delta$ (3); the policy finishes, and later rollouts reuse it (4). Orange marks what the person supplies; paths and actions are schematic.}
  \label{fig:overview}
\end{figure}

This paper asks two questions in order. When a generalist VLA fails under distribution shift, what kind of failure is it, and which kinds could the person watching it correct at all? And for the kind a person can reach, what is the narrowest correction a person can give, and is it enough? We answer the first with a taxonomy and the second with \corr{}, which grounds one sentence into an additive action bias on a frozen policy (Figure~\ref{fig:overview}). The bias need not steer the task, only bring the rollout back to where the policy can finish on its own (Section~\ref{sec:signal}).

\needspace{4\baselineskip}
\paragraph{Contributions.} We provide four findings in this paper.
\begin{enumerate}[leftmargin=*,itemsep=0pt,topsep=2pt]
  \item \textbf{A failure taxonomy that is stable across architectures.} On 133 failing instruction-model pairs, OpenVLA-OFT and $\pi_{0.5}$ fail in the same four ways in nearly the same proportions, and only in one mode, execution misalignment, is the error just a magnitude.
  \item \textbf{The task, not the episode, as the unit of correction.} One retrospective sentence per task is reused on every failed rollout in simulation and every trial of a robot run.
  \item \textbf{The boundary holds on hardware from both sides.} After an environment shift, one sentence lifts same-object pick-and-place from 10/30 to 30/30; two long-horizon tasks stay at 0/30.
  \item \textbf{The information is in the person.} A naively prompted GPT-5 mini, writing the same kind of correction from the same video, recovers no failure in up to 10 attempts, suggesting that what the person supplies is calibrated magnitude and timing.
\end{enumerate}

\section{Related Work}
\label{sec:related}

\textbf{Steering pretrained policies.} Fine-tuning a policy on new demonstrations~\citep{pi0,oft}, on language corrections such as ``move arm left''~\citep{rth,yay}, or on its own failures~\citep{afil} requires data and a redeployment for every shift, so a deployed robot needs a way to correct its frozen policy at test time. \emph{Steering} only selects or reshapes what the policy would produce, from its samples~\citep{itps,forewarn,wu2026dowhatyousay,zhang2026visual}, its prompt~\citep{jeong2026steering}, or its activations~\citep{haon2025mechanistic}; \emph{autonomous correctors} replan on visual deviation~\citep{vlacorrector}, steer with VLMs~\citep{grappa,vls}, or learn a chunk-correction head~\citep{a2c2}; and shared control needs a person in every episode~\citep{lilac,saps}. How frozen policies fail is less understood: runtime monitors~\citep{safe,sentinel,hideseek2026} say \emph{when} a rollout fails, and failure reasoners~\citep{reflect,aha} and analyses~\citep{paresmorlans2026demystifying} say why, but not which failures a person can correct. \corr{} starts from such an analysis, which shows why a person should engage: the correctable failures are errors of magnitude and timing, which a person easily recognizes but the policy repeats and autonomous correctors cannot yet judge. \textbf{Human corrections.} Corrections have been read as evidence about the objective~\citep{bajcsy2017physical,losey2018uncertainty,jeon2020reward}, also from their timing~\citep{wang2026timing}, turned into preferences for a reward~\citep{yang2024comparative,hirota2025active}, or followed as coaching by a policy trained for it~\citep{rth,yay,pi07}; all need a model that learns from them or was trained for them, though a correction there can also redirect the target or the plan. In \corr{}, for a frozen policy, one retrospective sentence per task, reused, becomes a bias on its actions, which every policy exposes, and can move a trajectory outside what the policy would produce, at the price of carrying no semantics.

\begin{figure}[t]
  \centering
  \includegraphics[width=\linewidth]{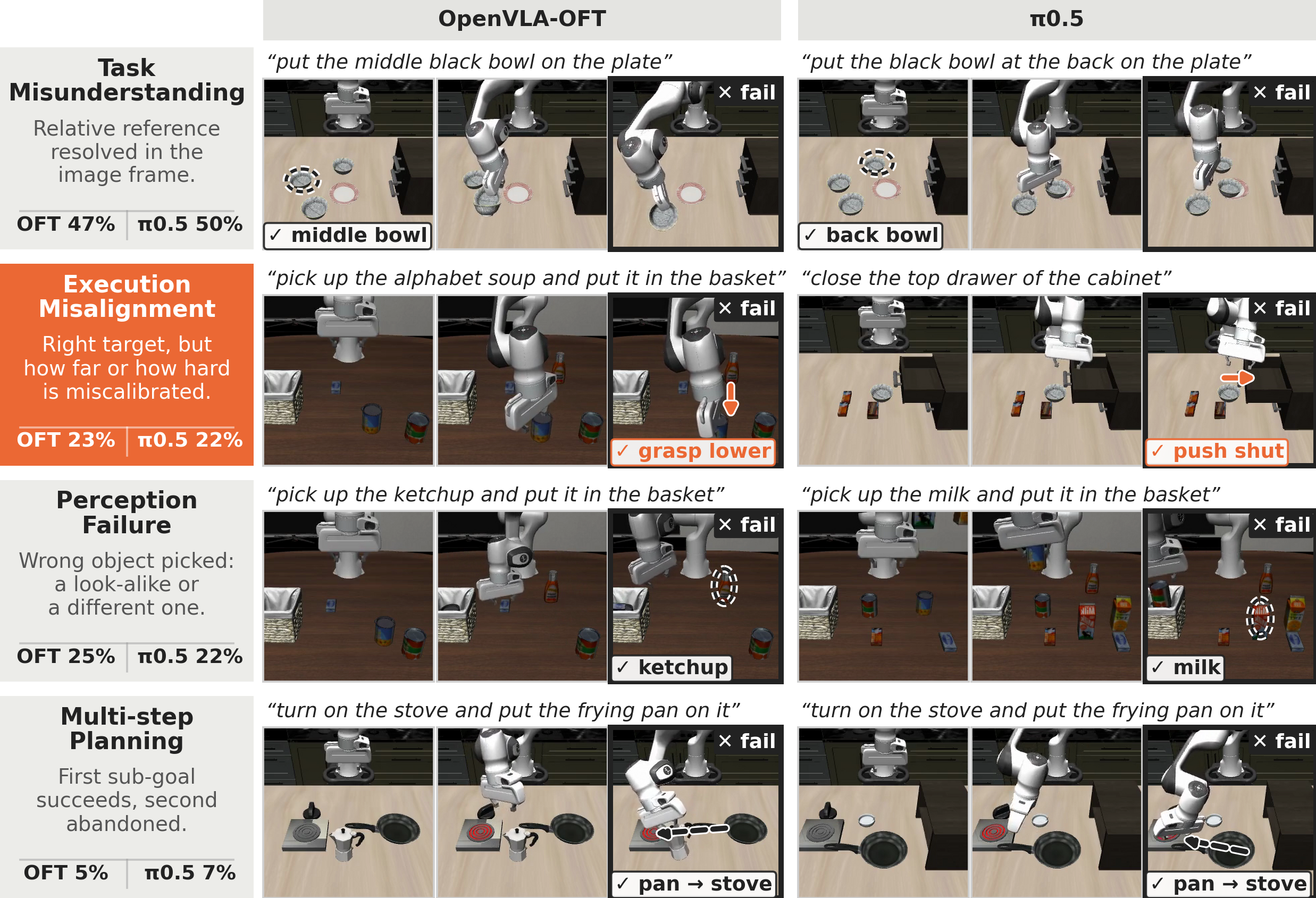}
  \caption{\textbf{Failure taxonomy.} Both VLAs show the same four modes in nearly the same proportions; in execution misalignment (orange), only the magnitude is wrong. \checkmark{} marks the correct answer.}
  \label{fig:taxonomy}
\end{figure}

\begin{figure}[t]
  \centering
  \includegraphics[width=\linewidth]{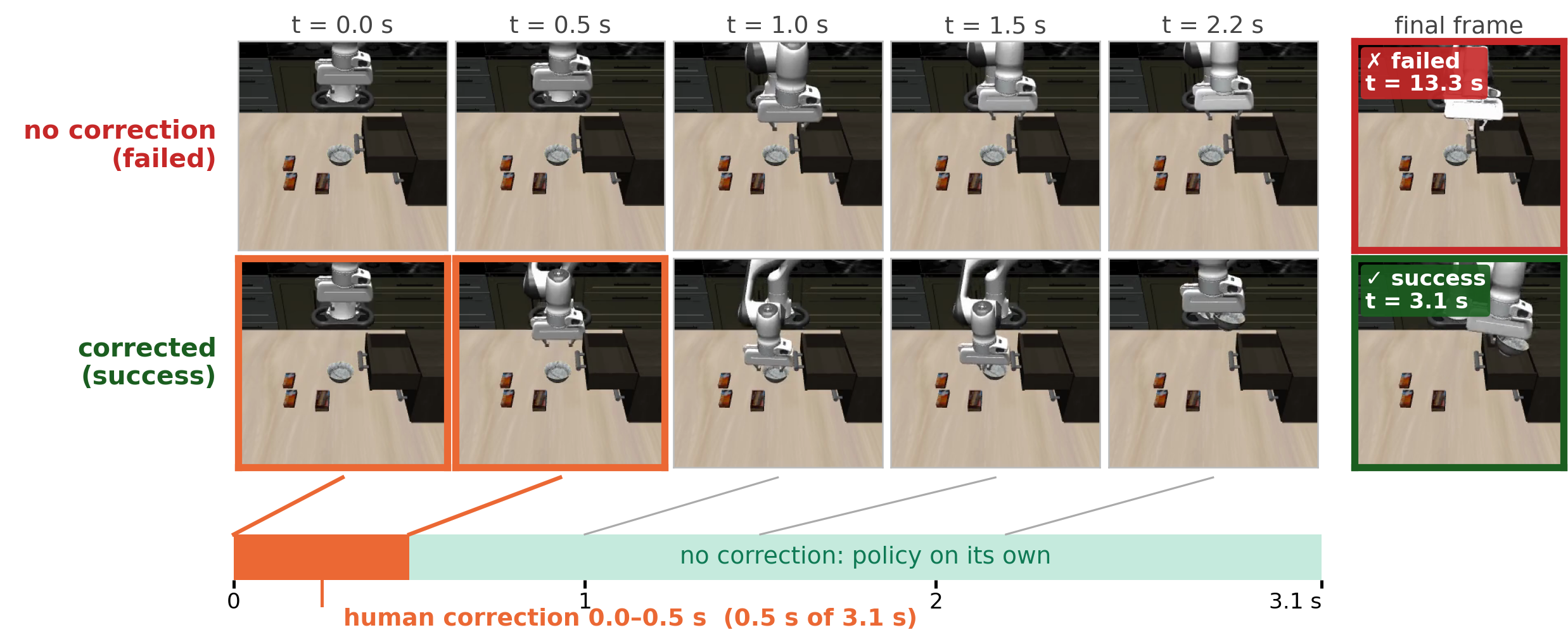}\\[4pt]
  \includegraphics[width=\linewidth]{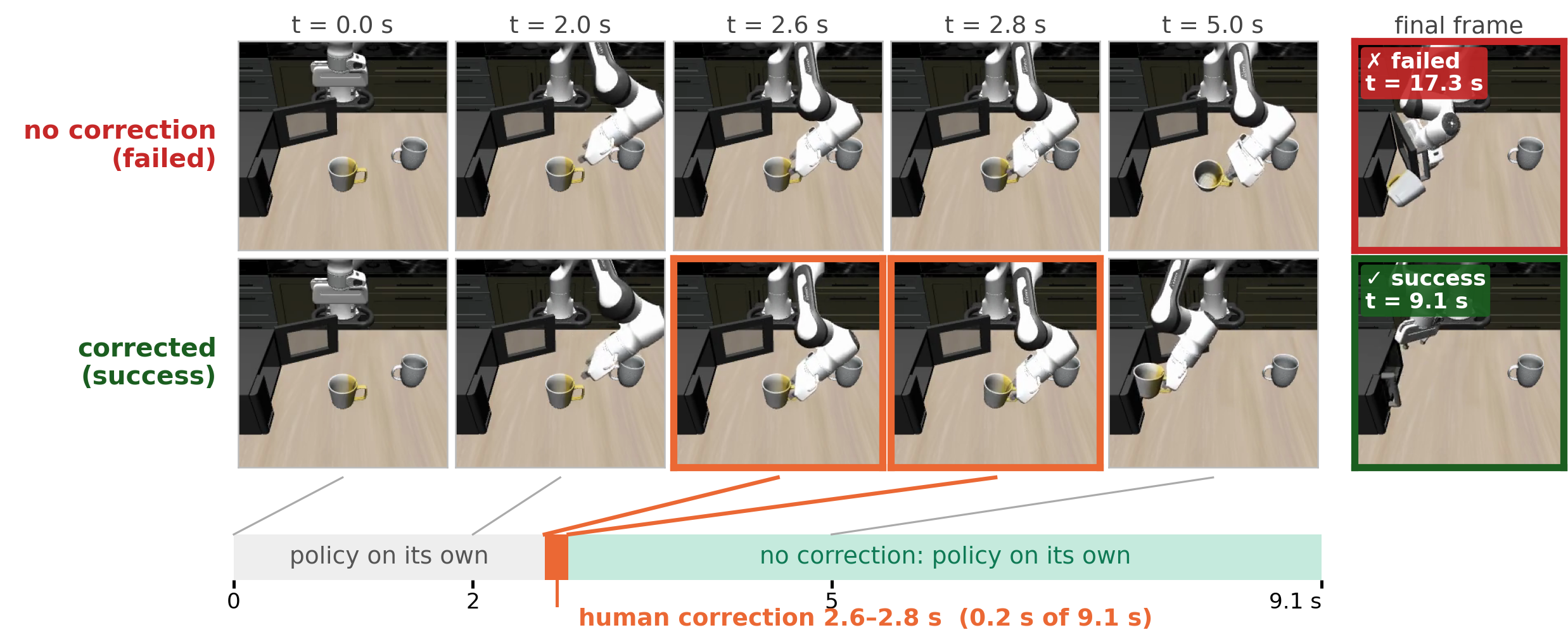}
  \caption{\textbf{One correction, then the policy on its own.} $\pi_{0.5}$; failed rollout above, corrected below, same times, plus each run's final frame; orange is the human's window. Top: LIBERO-90 (out of distribution), ``put the black bowl in the top drawer'', correction ``down, more, 0.0--0.5\,s''; uncorrected, it stays stuck at the top drawer until the 13.3\,s limit. Bottom: LIBERO-10 (in distribution), ``put the yellow and white mug in the microwave and close it'', correction ``forward a little, down a little, 2.6--2.8\,s''.}
  \label{fig:sparse}
\end{figure}

\section{Failure Taxonomy and Correctability}
\label{sec:taxonomy}

\paragraph{Protocol and four modes.}
We ran OpenVLA-OFT~\citep{oft} and $\pi_{0.5}$~\citep{pi05}, both fine-tuned on the same 40 LIBERO tasks (the Spatial, Object, Goal, and LIBERO-10 suites), on all 90 LIBERO-90 tasks~\citep{libero}, none of which is in that set. LIBERO-90 is therefore out of distribution in its instructions, not its scenes, which it shares with LIBERO-10. The 90 tasks carry 74 distinct instructions (12 recur across scenes); we label by instruction. Every instruction with at least one failed rollout was labelled by one annotator with the key failure frame and one of four modes, 73 of the 74 for OpenVLA-OFT and 60 for $\pi_{0.5}$ (the other 14 never failed), 133 instruction-model pairs. \emph{Task misunderstanding} (47\% of OpenVLA-OFT's failing instructions, 50\% of $\pi_{0.5}$'s): the policy resolves a relative spatial reference in the image frame rather than the table frame. Among identical bowls, ``the middle bowl'' (Figure~\ref{fig:taxonomy}, left) means the middle of the row on the table, but the policy takes the bowl in the middle of the image; ``the back bowl'' (right) means the bowl nearest the robot, but the policy takes the one on the left. That this is the largest mode fits evidence that VLAs trained on LIBERO often ignore the instruction~\citep{liberoplus} or misread a rephrasing~\citep{watts2026rephrase}. \emph{Perception failure} (25\%, 22\%): the policy picks the wrong object, either a look-alike or a plainly different one (asked for the milk, it takes the can). A wrong object counts as task misunderstanding when the instruction names it by a relative reference, and as perception failure when it names it directly. \emph{Execution misalignment} (23\%, 22\%): the right target is reached but how far or how hard is miscalibrated. \emph{Multi-step planning} (5\%, 7\%): the first sub-goal succeeds and the second is abandoned. Only execution misalignment keeps the right target and plan, so a magnitude bias should reach it and nothing else. That is this paper's operational boundary. Read the other way, it says what a robot should ask a person for: a magnitude in one mode, a target or a plan in the others.

\section{Human Correction: A Minimal Signal for the Correctable Mode}
\label{sec:signal}

\paragraph{Problem.}
A frozen VLA policy $\pi_\theta$ maps observations and an instruction to action chunks, executed one action $a_t\in\mathbb{R}^7$ ($x,y,z$, roll, pitch, yaw, gripper) at a time. A task $\tau$ fixes the instruction and the scene; its episodes start from random states $s_0$. A correction adds a bias, $a^{\mathrm{corr}}_t=a_t+\delta_t$: one sentence of $K$ clauses sets $\delta$ on one action dimension per clause, each in a short window, the latest ending at $t^{\mathrm{end}}$. With $\eta_\tau(\delta)$ the success rate under $\delta$, $\theta$ frozen, and no new data, we seek for each task one sentence whose $\delta$ improves it, $\Delta_\tau(\delta)=\eta_\tau(\delta)-\eta_\tau(0)>0$, as in language steering~\citep{jeong2026steering}. Let $V_\tau(s)$ be the chance that $\pi_\theta$ alone completes $\tau$ from $s$, a Markov state including queued actions, elapsed time, and a success flag. For a base and a corrected rollout from the same $s_0$ with fresh noise,
\begin{equation}
  \Delta_\tau(\delta) = p_\tau\,\rho_\tau(\delta)-(1-p_\tau)\,\beta_\tau(\delta) = \mathbb{E}\big[V_\tau(s_{t^{\mathrm{end}}})-V_\tau(s_0)\big],
  \label{eq:improvement}
\end{equation}
where $p_\tau$ is the base failure rate, $\rho_\tau$ the chance that the corrected rollout succeeds where the base failed, $\beta_\tau$ that it fails where the base succeeded, and $s_{t^{\mathrm{end}}}$ the corrected state when the bias ends, after which the policy acts alone. The gain is recovered failures minus broken successes, or the rise in the policy's own chance of finishing: a correction need only bring the state briefly to where $V_\tau$ is high, using what $\pi_\theta$ can already do. A task is \emph{correctable} if such a $\delta$ exists, which holds if (A1)~the error is one of magnitude, with target and plan right, so a short bias can raise $V_\tau$, and (A2)~a similar error, in size and time, recurs in every episode, so one $\delta$ serves them all. A person supplies $\delta$ if (A3)~they can read its direction, size, and timing from a failure video; comparative language conveys direction~\citep{yang2024comparative,hirota2025active}, and correction timing reveals intent~\citep{wang2026timing}.

\paragraph{From sentence to bias.}
A rollout fails; a person watches it once and writes one sentence. Each clause $f_k$ is grounded by GPT-5 mini into a 4-tuple
\begin{equation}
\begin{aligned}
  c_k &= \big(d_k,\; \sigma_k,\; m_k,\; [t^{\mathrm{start}}_k, t^{\mathrm{end}}_k]\big),\\
  d_k &\in \{x,y,z,\text{roll},\text{pitch},\text{yaw},\text{grip}\},\quad
  \sigma_k \in \{\pm1\},\quad
  m_k \in \{\text{slightly},\text{more},\text{much}\}.
\end{aligned}
\end{equation}
``Move down more in 1 to 2 seconds'' becomes $(z,-1,\text{more},[1.0,2.0])$. This is all the person supplies: explicit, retrospective, in end-effector space rather than a goal or preference, sparse ($K$ is one or two for most tasks, never more than five), and located in time. QuickLAP turns a similar parse, a feature mask and a preference shift, into reward weights~\citep{quicklap}; ours is applied as an action bias. Figure~\ref{fig:sparse} shows two examples: windows of 0.5\,s and 0.2\,s, and the policy alone for the rest. The term $m_k$ maps to a signed weight $w_k\in[-1,1]$, which the grounder picks inside a fixed range (\emph{slightly} $[0.3,0.5]$, \emph{more} $[0.5,0.8]$, \emph{much} $[0.8,1.0]$ of the controller's largest step, sign $\sigma_k$) set empirically from a few simulation trials; the person never types a size, and words such as ``a little'' map to the nearest term. The ranges depend on the robot and could be learned from comparative language feedback~\citep{yang2024comparative,hirota2025active}. A piecewise-linear envelope $\phi_k(t)$ ramps from 0 at $t^{\mathrm{start}}_k$ up to 1 and back to 0 at $t^{\mathrm{end}}_k$, and is zero outside the window. Clauses on the same dimension superpose:
\begin{equation}
  \delta_{t,d} = \sum_{k:\, d_k = d} w_k\,\phi_k(t).
  \label{eq:bias}
\end{equation}

\paragraph{Action-space adaptation.}
Corrections are specified in Cartesian space to match how people speak (``move left,'' ``move up''). The real xArm7 runs under joint-velocity control, so the Cartesian part $\dot{x}$ of $\delta_t$ is mapped to joint velocities through the Jacobian pseudoinverse~\citep{whitney1969resolved,siciliano2009robotics}, $\dot{q} = J^{\dagger}(q)\,\dot{x}$, where $J(q)$ is the manipulator Jacobian at the current configuration $q$; this is the minimum-norm joint velocity that realizes the requested end-effector motion.

\paragraph{Process.}
In Figure~\ref{fig:overview}'s steps, Diagnose keeps only execution misalignment (A1), where a sentence's expected gain can repay the person's time~\citep{dong2026voi}, and Correct gives $\delta$ by Eq.~(\ref{eq:bias}), revised if the corrected rollout still fails; most tasks succeeded with the first correction, and later rounds adjusted magnitude and direction. Simulation replays $\delta$ only after a failure and reports $(1-\hat p_\tau)+\hat p_\tau\hat\rho_\tau$, which counts chance recoveries and exceeds $\eta_\tau(\delta)$ by the unseen $(1-p_\tau)\beta_\tau$; replayed on all 50 microwave episodes (Figure~\ref{fig:sparse}), $\delta$ recovered 5 of 9 failures but lost 8 of 41 successes (net $-3$). The robot replays $\delta$ on every trial, measuring $\Delta_\tau$, which needs recoveries to outweigh breakages (A2).

\section{Does the Signal Suffice for Execution Misalignment?}
\label{sec:results}

\paragraph{Setup.}
Simulation uses LIBERO~\citep{libero} (\texttt{lerobot/\allowbreak pi05\_libero\_finetuned}~\citep{pi05}); the real robot is an xArm7 on RIO~\citep{rio} ($\pi_{0.5}$ fine-tuned from DROID~\citep{droid}). The \emph{VLM baseline} (GPT-5 mini, naive prompting, up to 10 refined attempts) writes the same 4-tuple from the failure video. Corrections are written only for tasks the same annotator labelled execution misalignment (12 in distribution).

\paragraph{Simulation.}
With 50 trials per LIBERO task in distribution and 5 per LIBERO-90 task out of distribution (5 per scene for a recurring instruction), a \corr{} retry raises success from 93.1\% to 96.3\% (65 of 139 failures recovered) and from 27.1\% to 34.4\% (33 of 328); the VLM baseline recovers none of these failures. Out of distribution the 33 are about half of the 64 failures in the 13 execution-misalignment instructions, though only four had corrections. Recovery was highest where a task failed similarly every time (A2), e.g., 9 of 10 failures on \emph{put the black bowl in the top drawer}, and lowest on \emph{close the top drawer}, which failed differently each time.

\begin{figure}[t]
  \centering
  \includegraphics[width=0.55\linewidth]{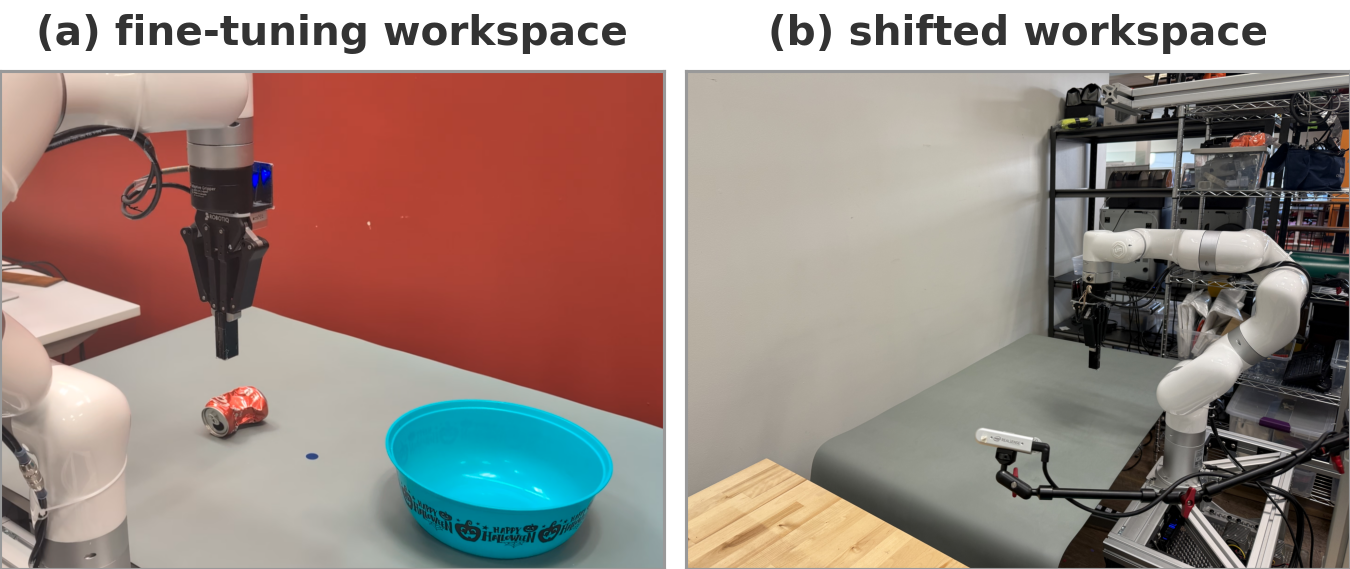}\hfill
  \includegraphics[width=0.43\linewidth]{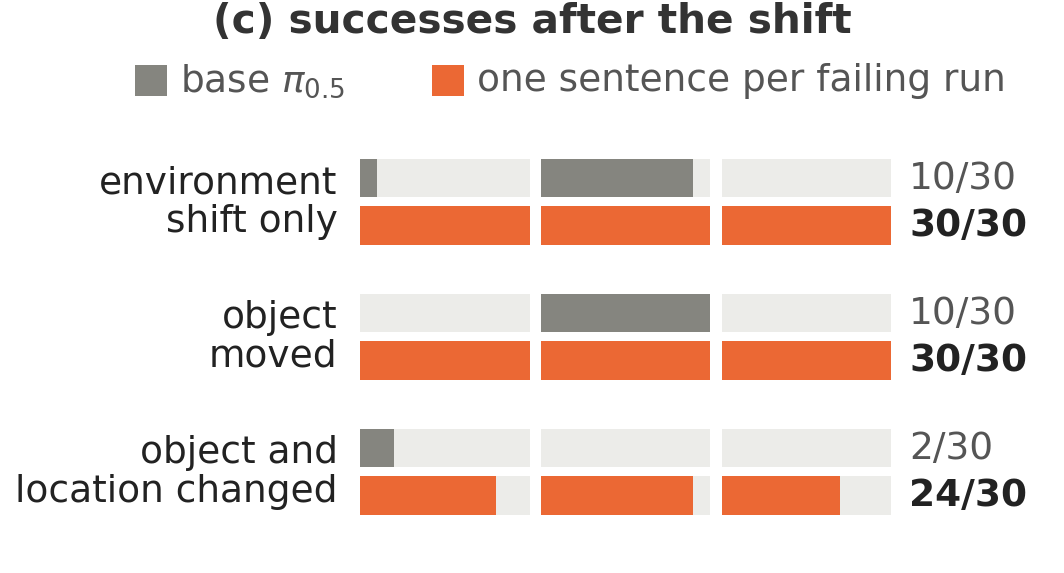}\\[3pt]
  \includegraphics[width=\linewidth]{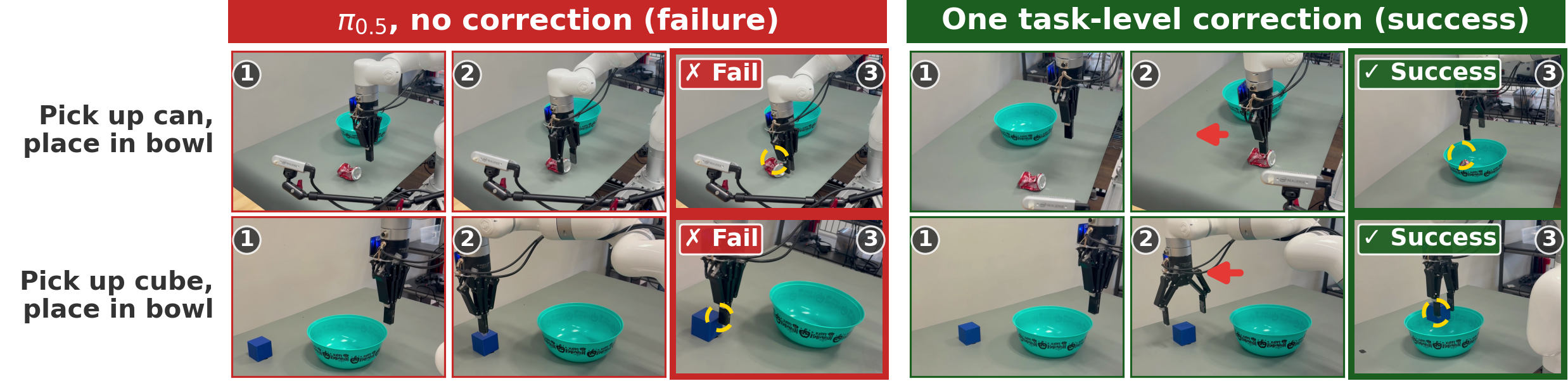}
  \caption{\textbf{Real robot.} (a)~Fine-tuning and (b)~shifted workspaces; (c)~successes after the shift, three runs of 10 trials (thirds). Bottom: after the shift $\pi_{0.5}$ misses both grasps (red); one task-level correction restores success (green).}
  \label{fig:real}
\end{figure}

\paragraph{Real robot.}
$\pi_{0.5}$-DROID reached 95.0\% on pick-and-place in its fine-tuning workspace~\citep{rio}. After we moved the robot base (Figure~\ref{fig:real}a,b), it scored 9/10 or 10/10 only in the run of each same-object condition whose object fell inside the fine-tuning distribution and 0/10 to 2/10 in the others, every failure a missed grasp, an execution misalignment (A1) that recurred within each run (A2). One correction per failing run, written in 40 to 80\,s and reused on all 10 trials, brings every run to 10/10 with the same object and to 7/10 to 9/10 with a new object (Figure~\ref{fig:real}c). Folding a shirt (a 92.5\% score before the shift~\citep{rio}) and stacking boxes fail in all 30 trials, with and without correction: the policy completes at most one step, a multi-step planning failure the sentence cannot reach.

\section{What Does the Human Supply?}
\label{sec:human}

Can feedback from a foundation model replace feedback from a person? For this channel, not yet. In our setup, the VLM baseline, given the same video and asked for the same 4-tuple, recovers nothing. The value seems to lie in two judgments the person made and the VLM did not (A3): how large the miss was in the robot's action units, and when it happened. Even though a stronger model does better on other tasks, it is unlikely to close this gap by itself: run as a policy on RoboDojo, GPT-6 Astra generalizes to semantic tasks but not to precise control, erring in lift height, approach angle or correction timing even when it appears to interpret the task correctly~\citep{astra2026}.

\section{Discussion: A Channel That Is Not Retained}
\label{sec:discussion}

\paragraph{The channel is misspecified for the failures that dominate.}
Over 70\% of failing instructions lack a target or a plan, and an additive bias cannot say ``the other bowl'' or fold a shirt. In the language of hypothesis-space misspecification~\citep{bobu2020misspecification}, the channel cannot represent what the person means, so no feedback converges. Yet a frontier model run as a policy is strongest on semantic tasks~\citep{astra2026}.

\paragraph{Nothing is learned.}
The correction is written by a person and replayed as given; every new shift or task needs a new sentence. The open problem is \emph{intent propagation}: can a policy retain what a correction expresses and generalize it beyond the situation in which it was said? Recent work infers reward structure from language and demonstrations~\citep{maskedirl} or corrections~\citep{quicklap}, generalizes intent to new states at test time~\citep{gift}, and compiles local corrections into reusable knowledge and skills~\citep{droc,memo,architect,vocalsandbox}.

\section{Limitations and Conclusion}
\label{sec:conclusion}

\paragraph{Limitations.}
The taxonomy was labelled by one annotator; there was no window sweep and no constant-offset, plain-retry, or structured-interface baseline; simulation retried only failed rollouts; corrections were written by the authors, with no user study; magnitude ranges were set empirically and not learned.

\paragraph{Conclusion.}
Under new instructions, a generalist VLA fails in four ways, and only execution misalignment errs in magnitude alone. For that mode the smallest correction a person can give, one retrospective sentence per task, is often enough: added as a bias to the frozen policy, it brings the rollout back to where the policy can finish on its own and recovers failures in simulation and on a real robot after an environment shift, and one VLM given the same video did not. Two lessons follow: diagnose the mode before choosing the channel, and treat the sentence as information a policy should keep; today it is replayed, never learned.

\clearpage
\bibliography{refs}

\end{document}